\documentclass{article}

\PassOptionsToPackage{numbers}{natbib}
\usepackage[final]{neurips_2023_gaied}

\usepackage[utf8]{inputenc} 
\usepackage[T1]{fontenc}    
\usepackage{hyperref}       
\usepackage{url}            
\usepackage{booktabs}       
\usepackage{amsfonts}       
\usepackage{nicefrac}       
\usepackage{microtype}      
\usepackage{xcolor}         

\usepackage{graphicx}
\usepackage{caption}
\usepackage{subcaption}

\title{Retrieval-augmented Generation to Improve Math Question-Answering: Trade-offs Between Groundedness and Human Preference}

\author{%
Zachary Levonian$^1$ \quad Chenglu Li$^2$ \quad Wangda Zhu$^3$ \quad Anoushka Gade$^1$ \\
\textbf{Owen Henkel}$^4$ \quad \textbf{Millie-Ellen Postle}$^5$ \quad \textbf{Wanli Xing}$^3$\\
$^1$Digital Harbor Foundation \quad $^2$University of Utah \quad $^3$University of Florida \\ 
$^4$University of Oxford \quad $^5$Rising Academies \\
\texttt{zach@digitalharbor.org}
}

\begin{document}

\maketitle

\begin{abstract}
For middle-school math students, interactive question-answering (QA) with tutors is an effective way to learn.
The flexibility and emergent capabilities of generative large language models (LLMs) has led to a surge of interest in automating portions of the tutoring process---including interactive QA to support conceptual discussion of mathematical concepts.
However, LLM responses to math questions can be incorrect or mismatched to the educational context---such as being misaligned with a school's curriculum.
One potential solution is retrieval-augmented generation (RAG), which involves incorporating a vetted external knowledge source in the LLM prompt to increase response quality.
In this paper, we designed prompts that retrieve and use content from a high-quality open-source math textbook to generate responses to real student questions. 
We evaluate the efficacy of this RAG system for middle-school algebra and geometry QA by administering a multi-condition survey, finding that humans prefer responses generated using RAG, but not when responses are \textit{too} grounded in the textbook content.
We argue that while RAG is able to improve response quality, designers of math QA systems must consider trade-offs between generating responses preferred by students and responses closely matched to specific educational resources. 


\end{abstract}


\section{Introduction}



According to the National Assessment of Educational Progress (NAEP), nearly 40\% of high school students lack a basic grasp of mathematical concepts~\cite{naepNAEPMathematicsNational}, underscoring the need to enhance math education in K-12 environments.
One of the most impactful methods to support students' math learning is through math question and answer (QA) sessions tutored by humans. 
Math QA can be approached with two main focuses: (1) enhancing students' \textit{procedural} fluency with strategies such as step-by-step problem solving for specific math topics and (2) deepening students' \textit{conceptual} understanding through scaffolding such as clarifying math concepts with concrete or worked examples, providing immediate feedback, and connecting math ideas to real-world scenarios~\cite{moschkovichScaffoldingStudentParticipation2015, rittle-johnsonNotOnewayStreet2015, hurrellConceptualKnowledgeProcedural2021}. While tutor-led math QA is effective~\cite{nickow_impressive_2020}, they face challenges such as efficiently allocating tutoring resources, ensuring wide accessibility due to high costs, and scaling up to support a myriad of learners with consistent quality ~\cite{kraftBlueprintScalingTutoring2021, cukurovaLearningAnalyticsApproach2022}. 

To address these challenges in math QA, educational researchers have sought AI to build expert systems and intelligent tutoring systems to enhance math learning with procedural practice~\cite{ritterCognitiveTutorApplied2007, arroyoUsingIntelligentTutor2011,aleven_towards_2023}. 
However, limited educational research has focused on the potential of AI for improving students' conceptual understanding of math concepts. This study is a preliminary attempt to fill that gap by building the understanding needed to deploy conceptual math QA.
We formed a research partnership with the developers of Rori, a WhatsApp-based chatbot math tutor primarily used by low-income middle-school students in Sierra Leone, Liberia, Ghana, and Rwanda.\footnote{\url{https://rori.ai}}.
While Rori uses a chat interface, its pedagogical approach is based on intelligent tutoring systems (ITS) and it adopts a mastery-based learning approach that takes students through procedural lessons based on their abilities. 
Rori is currently designing for the inclusion of conceptual math QA using LLMs.
There have been preliminary efforts to use LLMs in educational settings to scaffold student discussions, to provide feedback~\cite{kasneci_chatgpt_2023}, to personalize learning experiences through automatic text analysis and generative socio-emotional support~\cite{sungHowDoesAugmented2021, liNaturalLanguageGeneration2021}, and to extend LLMs for many other educational tasks~\cite{shenMathBERTPretrainedLanguage2021}.


While the results from these education-related LLM explorations are encouraging, there are ethical considerations when using LLM outputs for math education~\cite{kasneci_chatgpt_2023,nye_generative_2023}. 
A primary concern is \textit{hallucinations}, where LLMs generate answers that sound plausible and coherent but are factually incorrect~\cite{dziri_faithdial_2022}.
Such misleading yet persuasive responses from LLMs could inadvertently instill incorrect conceptual understanding in students.
Researchers from the AI community have investigated strategies to mitigate LLM hallucinations (see \citeauthor{jiSurveyHallucinationNatural2023}'s review \cite{jiSurveyHallucinationNatural2023}), with retrieval-augmented generation (RAG) standing out given its effectiveness and flexibility of implementation (e.g., model agnostic)~\cite{lewis_retrieval-augmented_2020,yang_leandojo_2023}. 
Conceptually, RAG in an educational context aims to bolster the correctness of LLM-based QA by drawing from external knowledge sources such as syllabi, workbooks, and handouts, such that the LLM's responses are, to various extents, anchored to established learning materials~\cite{peng_check_2023}.
An interactive student chat backed by RAG offers the promise of both high correctness and faithfulness to materials in a vetted curriculum.
Grounding tutoring materials in a student's particular educational context is an important requirement for system adoption~\cite{yang_can_2021,holstein_intelligent_2017}.

\begin{figure}
  \centering
  \includegraphics[width=\textwidth]{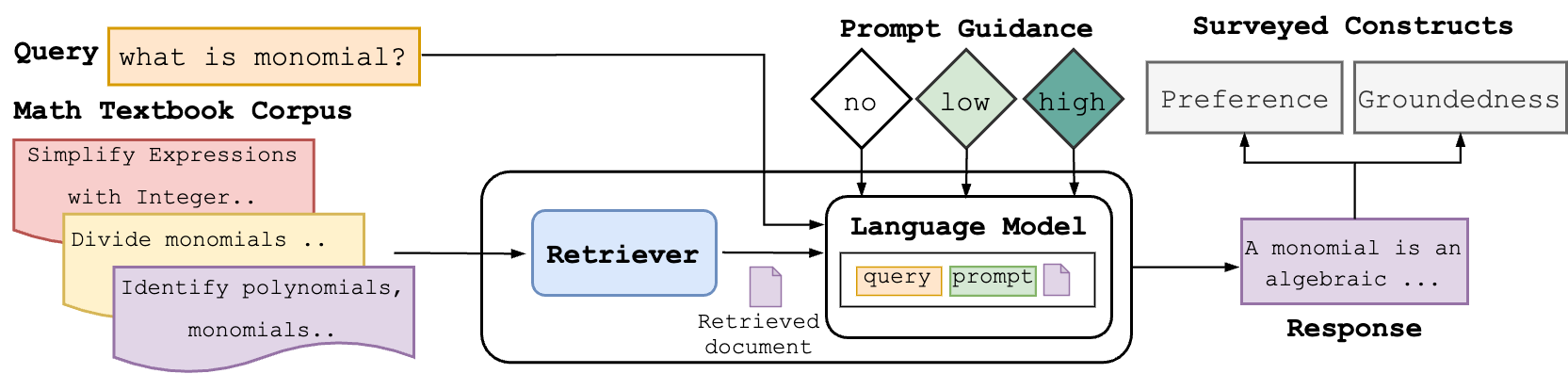}
  \caption{In this paper, we generated responses to math student queries with a retrieval-augmented generation system using one of three prompt guidance conditions. Survey respondents ranked responses by preference and assessed groundedness in the underlying math textbook used as a retrieval corpus.}
  \label{fig:overview}
\end{figure}

We implemented a RAG system for conceptual math QA (described in sec.~\ref{sec:system}).
To evaluate our RAG system, we started with the problem of designing prompts that produce both the expected tutor-like behavior and responses grounded in the retrieved document.
Can we use retrieval-augmented generation and prompt engineering to increase the groundedness of LLM responses?
In study 1 (sec.~\ref{sec:study1}), we observe qualitative trade-offs in response quality and the level of guidance provided in the LLM prompt, motivating quantitative study of human preferences.
Do humans prefer more grounded responses?
In study 2 (sec.~\ref{sec:study2}), we survey preferences for LLM responses at three different levels of prompted guidance, finding that the most-preferred responses strike a balance between no guidance and high guidance.
How does retrieval relevance affect response groundedness?
In study 3 (sec.~\ref{sec:study3}), we consider the impact of document relevance on observed preferences.
Fig.~\ref{fig:overview} shows an overview of the RAG system and its use for addressing our research questions.

\section{Related Work}
\label{sec:background}

Intelligent Tutoring Systems (ITSs)
are educational technologies designed to provide one-on-one instructional guidance  comparable to that of expert human tutors~\cite{psotka_intelligent_1988}. Structurally, ITSs implement a user interface over a knowledge base with a pedagogical model that determines how the ITS should respond to student inputs~\cite{sedlmeier_intelligent_2001}. ITSs are traditionally based on iteratively serving procedural lesson content and providing hints in response to student mistakes~\cite{vanlehn_behavior_2006}.
ITSs have been shown to be effective as tutors in specific domains such as mathematics and physics~\cite{vanlehn_relative_2011}.
To extend an ITS that currently focuses on procedural fluency with features focused on conceptual understanding~\cite{sottilare_design_2014}, we turn to the flexibility and expressive power of LLMs. 
LLMs have been proposed as useful for supporting a large number of education-related tasks~\cite{caines_application_2023,kasneci_chatgpt_2023}. 
Despite the potential utility of LLMs for education, there are significant concerns around their correctness and ability to meet students at their appropriate level~\cite{kasneci_chatgpt_2023}.
LLMs have been used in procedural tutoring and problem-solving systems, with careful prompt engineering used to improve reliability~\cite{upadhyay_improving_2023}.
A more complex approach is using retrieval to augment the LLM prompt in order to improve response quality.
For example, the SPOCK system for biology education retrieves relevant textbook snippets when generating hints or providing feedback~\cite{sonkar_class_2023}.
Retrieval-augmented generation (RAG) involves retrieving texts from an external corpus relevant to the task and making them available to the LLM~\cite{lewis_retrieval-augmented_2020,peng_check_2023}.
RAG has been used to improve diverse task performance of LLMs~\cite{mialon_augmented_2023}, either by incorporating retrieved texts via cross-attention~\cite{izacard_atlas_2022,borgeaud_improving_2022,lewis_retrieval-augmented_2020} or by inserting retrieved documents directly in the prompt~\cite{guu_realm_2020}.\footnote{A note on terminology: in \citeauthor{lewis_retrieval-augmented_2020}’s paper proposing ``retrieval-augmented generation'', they used the term to refer to an underlying LLM trained or fine-tuned with retrieved documents. The term has come to refer to any combination of LLMs and document retrieval: the method we use in this paper follows the common approach of using in-context learning rather than fine-tuning~\cite{lazaridou_internet-augmented_2022,lu_are_2023}. A better term for these approaches may be ``retrieval-enhanced machine learning''~\cite{zamani_retrieval-enhanced_2022}, which includes pre-LLM neural models using retrieval e.g.~\cite{chen_reading_2017}.}
We apply RAG in the education domain by using a math textbook as an external corpus and evaluating if RAG leads to responses that are preferred more often by humans and grounded in the textbook content.

\section{A retrieval-augmented generation system for math question-answering}
\label{sec:system}

To support the development of reliable conceptual question-answering in a math chatbot, we implemented a retrieval-augmented generation system backed by a vetted corpora of math content, e.g. lesson plans, textbooks, and worked examples.
RAG cannot provide a benefit during generation if the retrieved documents are not relevant, so we intentionally selected a corpus that will be relevant to many math-related student questions but not to all plausible questions.

\textbf{OpenStax Prealgebra retrieval corpus}
We selected a Prealgebra textbook made available by OpenStax~\cite{marecek_prealgebra_2020}, segmented by sub-section. The textbook covers whole numbers, functions, and geometry, among other topics.

\textbf{RAG implementation}
We adopted a commercially-realistic chatbot context as the underlying LLM, generating all responses with the OpenAI API using model gpt-3.5-turbo-0613 with default temperature settings.
We built on our own implementation of RAG~\cite{levonian_llm-math-education_2023} that uses a variant of parent retrieval~\cite{chase_parent_2023}.
When a student asks a question, we identify a single relevant section of the textbook using cosine similarity against dense representations of the query and the textbook subsections.
We created all representations using OpenAI's text-embedding-ada-002 model~\cite{greene_new_2022},
an effective dense text embedding model~\cite{muennighoff_mteb_2023}.
We released our code and data on GitHub.\footnote{\url{https://github.com/DigitalHarborFoundation/rag-for-math-qa}}
Additional details in App.~\ref{app:sec:implementation}.

\section{Study 1: Can we use retrieval-augmented generation and prompt engineering to increase the groundedness of LLM responses?}
\label{sec:study1}

\begin{table}[]
\caption{Representative student questions in the 51 Math Nation queries.}
\centering
\begin{tabular}{@{}l|l@{}}
\toprule
Can I get the steps for factoring quadratics & What is the domain and range? How do I find it? \\
How do I add line segments again?? & How do you know if a number is a constant? \\
what is monomial & How do I multiply fractions??????? \\ \bottomrule
\end{tabular}
\label{tab:query_examples}
\end{table}

In using RAG, we hope that system responses will both answer the student's query and reflect the contents of the retrieved document.
As the retrieved document cannot be perfectly relevant for all queries, achieving this \textit{groundedness} may require producing inaccurate or otherwise less useful responses.
Thus, there is an apparent trade-off between groundedness and the perceived usefulness of the system response.
If this trade-off exists, we may want to influence the balance between groundedness and usefulness by adjusting the system prompt.
This first study tackles a basic question: \textit{can} we influence this balance by engineering the prompt?
We now introduce the prompt guidance conditions we used, the queries used for evaluation, and three evaluation metrics.

\textbf{Guidance conditions}
Prompt engineering is important for LLM performance~\cite{mishra_reframing_2022,lu_are_2023,upadhyay_improving_2023}.
Each guidance condition was selected by iterative, qualitative exploration of prompts given 1-3 sample student questions. 
While these prompts are unlikely to be ``optimal''~\cite{yang_large_2023}, they produce reasonable outputs. 
The \textbf{No guidance} condition does not use RAG and contains a simple prompt that begins: ``You are going to act as a mathematics tutor for a 13 year old student who is in grade 8 or 9 and lives in Ghana. You will be encouraging and factual. Prefer simple, short responses.'' 
Other prompts build on this basic instruction set---see App.~\ref{app:sec:prompts}. 
The \textbf{Low guidance} prompt adds ``Only if it is relevant, examples and language from the section below may be helpful to format your response:'' followed by the retrieved document.
The \textbf{High guidance} prompt instead says ``Reference content from this textbook section in your response:''.
The \textbf{Information Retrieval} condition---used only in this first study to demonstrate the shortfalls of automated metrics for conversational responses---says ``Repeat the student's question and then repeat in full the most relevant paragraph from my math textbook.''

\textbf{Student queries}
Math Nation is an online math platform with an interactive discussion board~\cite{banawan_math_2022}.
On this board, students seek help on math-related questions supported by their instructors, paid tutors, and peers.
We annotated a random sample of 554 Math Nation posts made by students between October 2013 and October 2021 on boards for Pre-algebra, Algebra 1, and Geometry.
We identified 51 factual and conceptual questions that have sufficient context to be answerable; the majority of excluded questions sought procedural help.
Representative questions are shown in Table~\ref{tab:query_examples}.

\textbf{Evaluation metrics}
Given the relative novelty of our task, automatically measuring usefulness or correctness is not feasible.
However, there is a large body of information retrieval (IR) literature on measuring groundedness of a generated text.
We adopt three metrics used in prior work~\cite{adlakha_evaluating_2023,chiesurin_dangers_2023,dziri_faithdial_2022,rajpurkar_squad_2016}.
K-F1++ is a token-level metric that completely ignores semantics, proposed by \citeauthor{chiesurin_dangers_2023} as more appropriate for conversational QA than Knowledge F1~\cite{chiesurin_dangers_2023}.
BERTScore is a token-level metric that uses RoBERTa-base embeddings to model semantics~\cite{zhang_bertscore_2020}.
BLEURT is a passage-level metric that models semantics using BERT-base fine-tuned on human relevance judgments~\cite{sellam_bleurt_2020}.

\begin{figure}
  \centering
  \includegraphics[width=\textwidth]{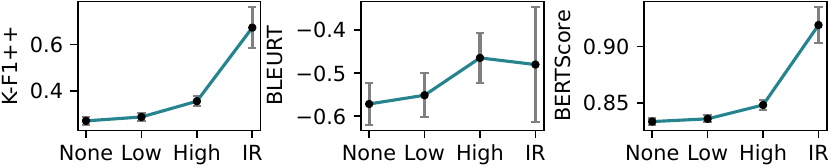}
  \caption{Groundedness for four levels of prompt guidance.}
  \label{fig:automated_metrics}
\end{figure}

\textbf{Results}
Fig.~\ref{fig:automated_metrics} shows that metric values on the 51 queries increase across guidance conditions.
All confidence intervals are computed at the 95\% significance level.
These results confirm our basic intuition that groundedness is manipulable with prompt engineering.
We do not know if response quality stays the same, increases, or even decreases as groundedness increases, but the results of the IR condition suggest that it \textit{might} decrease: while the token-level metrics indicate that IR is the most grounded condition, its responses include no conversational adaptation to the student's question and so are lower quality in our context.
In study 2, we will directly address the questions of response quality and groundedness by surveying humans.


\section{Study 2: Do humans prefer more grounded responses?}
\label{sec:study2}

\begin{figure}
  \centering
  \begin{subfigure}[b]{0.5\textwidth}
     \centering
     \includegraphics[width=\textwidth]{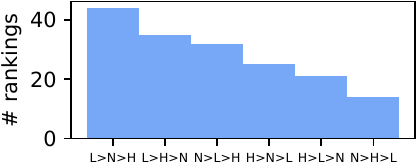}
     \caption{Distribution of all 144 preference rankings.}
     \label{fig:rank_distribution}
 \end{subfigure}
 \hfill
 \begin{subfigure}[b]{0.49\textwidth}
     \centering
     \includegraphics[width=\textwidth]{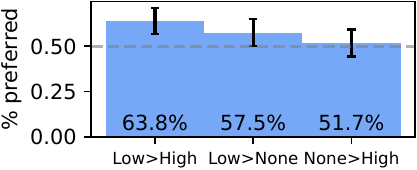}
     \caption{Pair-wise preferences from all rankings.}
     \label{fig:pairwise_ranks}
 \end{subfigure}
  \caption{Ranked preferences for LLM responses in three guidance conditions: no guidance (N), low guidance (L), and high guidance (H).}
  \label{fig:rank_preference}
\end{figure}

\textbf{Methods}
To understand the impact of guidance on human preference for LLM responses, we surveyed 9 educators and designers of education technologies.
We selected a comparative (within-subjects) design: with query and response order randomized, respondents ranked from best to worst the responses generated in the None, Low, and High guidance conditions for each query.
To determine if the guidance conditions were perceived to be grounded in the retrieved document, we adapted a scale used in prior work as an ordinal None (0), Partial (1), Perfect (2) judgment~\cite{adlakha_evaluating_2023}.
Responses were spread across four Qualtrics surveys; all questions received 3-4 responses. The survey is in App.~\ref{app:sec:ranking_survey}.

\begin{figure}
  \centering
  \begin{subfigure}[b]{0.5\textwidth}
     \centering
     \includegraphics[width=\textwidth]{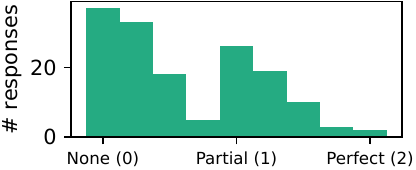}
     \caption{Distribution of groundedness across responses in all three guidance conditions, averaged over annotators.}
     \label{fig:mean_faithfulness_distribution}
 \end{subfigure}
 \hfill
 \begin{subfigure}[b]{0.49\textwidth}
     \centering
     \includegraphics[width=\textwidth]{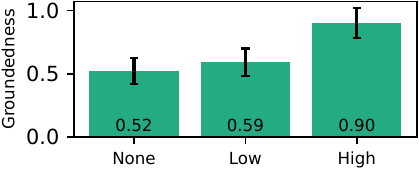}
     \caption{Mean groundedness by guidance condition across all queries and annotators.}
     \label{fig:faithfulness_by_guidance}
 \end{subfigure}
  \caption{Groundedness of the generated responses on an ordinal None (0), Partial (1), Perfect (2) scale.}
  \label{fig:faithfulness}
\end{figure}

\textbf{Results}
Fig.~\ref{fig:rank_preference} shows respondent preferences for the three guidance conditions.
Responses in the low guidance condition are preferred over responses in the no guidance \textit{and} high guidance conditions.
The high and no guidance conditions were statistically indistinguishable.
At least two of the guidance conditions significantly differ in groundedness ($n$=153, one-way ANOVA F(2.0, 99.38)=6.65, p=0.001).
We observed substantial inter-rater variation for groundedness ($n=153$, Krippendorff's $\alpha$=0.35).
Fig.~\ref{fig:faithfulness} shows that respondents do perceive high guidance responses to be more grounded in the retrieved document than low and no guidance responses.
Surprisingly, low guidance responses are not perceived to be significantly more grounded than no guidance responses, suggesting that low guidance responses are preferred for reasons other than their groundedness, a question we will investigate further in study 3.\footnote{Notably, there is no meaningful correlation between the rank of a low guidance response and its perceived faithfulness (Pearson's $r$=-0.08, p=0.29).}

\section{Study 3: How does retrieval relevance affect response groundedness?}
\label{sec:study3}

\textbf{Methods} 
It may be that responses in the low guidance condition were preferred by survey respondents because the LLM includes content in the retrieved document if it is relevant and omits it if not.
To test this hypothesis, three of the authors independently annotated each query and the associated retrieved document for relevance using a four-point ordinal scale used in prior work~\cite{hofstatter_fine-grained_2020,althammer_tripjudge_2022}---see App.~\ref{app:sec:relevance_survey}.

\begin{figure}
  \centering
  \begin{subfigure}[b]{0.5\textwidth}
     \centering
     \includegraphics[width=\textwidth]{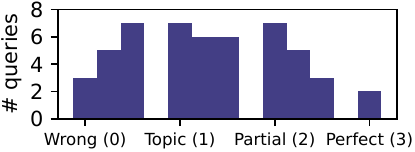}
     \caption{Distribution of retrieved document relevance for all queries, averaged over annotators.}
     \label{fig:mean_relevance_distribution}
 \end{subfigure}
 \hfill
 \begin{subfigure}[b]{0.49\textwidth}
     \centering
     \includegraphics[width=\textwidth]{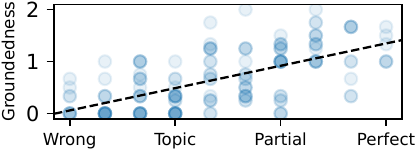}
     \caption{Correlation between perceived groundedness and retrieved document relevance ($r$=0.56, p<0.001).}
     \label{fig:relevance_faithfulness}
 \end{subfigure}
  \caption{Human-annotated relevance of the retrieved document for all 51 queries.}
  \label{fig:relevance}
\end{figure}

\textbf{Results}
Inter-rater reliability was generally low ($n=51$, Fleiss' $\kappa = 0.13$, Krippendorff's $\alpha = 0.40$).
For subsequent analysis, we computed the mean relevance of each document across annotators.
70.6\% of queries are deemed at least topically relevant, while 33.3\% are deemed partially relevant or better; see Fig.~\ref{fig:mean_relevance_distribution} for the full distribution.
Across all guidance conditions, responses were more likely to be grounded if the retrieved document is relevant (Fig.~\ref{fig:relevance_faithfulness}).
However, we observed no significant relationship between relevance and preference (rank).
For example, for queries where low guidance responses are preferred over high guidance responses, mean relevance is actually slightly \textit{higher} (diff=0.19, $t$=-1.45, p=0.15).

\textbf{Correlation between human annotations and automated metrics}
Given the results in study 2 suggesting that low guidance responses are not perceived to be more grounded than no guidance responses, we were further interested in possible correlations between perceived groundedness or relevance and the automated groundedness metrics.
Table \ref{tab:human_vs_automated} shows modest positive correlations between automated groundedness metrics and human annotations. 
K-F1++ has the strongest correlation ($r$=0.52) with groundedness, although the correlation is weaker as guidance decreases.

\begin{table}
\caption{Correlation between human annotations and automated groundedness metrics. Pearson's $r$ with p-values Bonferroni-corrected for 12 comparisons. Note: $^{*}$p<0.05, $^{**}$p<0.01, $^{***}$p<0.001.}
\centering
\begin{tabular}{@{}lllllll@{}}
\toprule
Guidance & \multicolumn{3}{c}{Faithfulness} & \multicolumn{3}{c}{Relevance} \\
& K-F1++   & BLEURT   & BERTScore  & K-F1++  & BLEURT  & BERTScore \\ \midrule
None & 0.38&0.33&0.35&0.26&0.34&0.43$^{*}$ \\
Low & 0.47$^{**}$&0.32&0.61$^{***}$&0.43$^{*}$&0.34&0.50$^{**}$ \\
High & 0.50$^{**}$&0.21&0.39&0.37&0.26&0.50$^{**}$ \\ \midrule
Pooled & 0.52$^{***}$&0.33$^{***}$&0.51$^{***}$&0.31$^{**}$&0.30$^{**}$&0.42$^{***}$ \\ \bottomrule
\end{tabular}
\label{tab:human_vs_automated}
\end{table}

\section{Implications \& Future Work}

Across three studies, we investigated prompt engineering as a guidance mechanism alongside retrieval-augmented generation to encourage high-quality and grounded responses that are appropriate for students.
Our most important finding is that \textbf{humans prefer responses to conceptual math questions when retrieval-augmented generation is used, but only if the prompt is not ``too guiding''.}
While RAG is able to improve response quality, we argue that designers of math QA systems should consider trade-offs between generating responses preferred by humans and responses closely matched to specific educational resources.
Math QA systems exist within a broader socio-technical educational context; the pedagogically optimal response may not be the one preferred by the student at that time.
\citeauthor{chiesurin_dangers_2023} distinguish between groundedness---when a response is found in the retrieved document---and \textit{faithfulness}---when the response is both grounded and answers the query effectively~\cite{chiesurin_dangers_2023}.
Faithfulness is a desirable property for conceptual math QA systems, and we view designing for and evaluating faithfulness as an open problem.
Our results show that prompt guidance with RAG is one potential design knob to navigate faithfulness.
Future work might improve understanding of faithfulness by building taxonomies based on educational theories of effective tutoring, adapting existing procedural faithfulness metrics (e.g.,~\cite{adlakha_evaluating_2023,dziri_faithdial_2022}), and explaining the role of retrieved document relevance (as in our surprising study 3 results finding that relevance was not a meaningful predictor of human preference).

This paper is a preliminary step toward understanding the relationship between groundedness and preference in conceptual math QA systems.
Future work must extend beyond single-turn responses to include exploration of follow-up questions~\cite{wang_mint_2023} and to design for the actual context of use.
The most important limitation of this work is that we did not collect preferences directly from middle-school students, although we did use real student questions.
Qualitative research of students' preferences should focus not only on correctness but also on factors such as conceptual granularity, curriculuar alignment, and cultural relevance.
We were concerned about the ethics of presenting an untested math QA system to students but are now combining insights from these results with the implementation of guard-rails to deploy a safe in-classroom study.
Beyond preferences, future math QA systems that use RAG will need to explore the relationship between students' response preferences and actual learning outcomes.

\begin{ack}
We would like to thank Bill Roberts, Ralph Abboud, and the staff of Rising Academies for their contributions.
This work was supported by the Learning Engineering Virtual Institute (LEVI) and funded by the Digital Harbor Foundation.
\end{ack}

\bibliographystyle{ACM-Reference-Format}
\bibliography{enghub_bibtex, chenglu}


\begin{thebibliography}{56}


\ifx \showCODEN    \undefined \def \showCODEN     #1{\unskip}     \fi
\ifx \showDOI      \undefined \def \showDOI       #1{#1}\fi
\ifx \showISBNx    \undefined \def \showISBNx     #1{\unskip}     \fi
\ifx \showISBNxiii \undefined \def \showISBNxiii  #1{\unskip}     \fi
\ifx \showISSN     \undefined \def \showISSN      #1{\unskip}     \fi
\ifx \showLCCN     \undefined \def \showLCCN      #1{\unskip}     \fi
\ifx \shownote     \undefined \def \shownote      #1{#1}          \fi
\ifx \showarticletitle \undefined \def \showarticletitle #1{#1}   \fi
\ifx \showURL      \undefined \def \showURL       {\relax}        \fi
\providecommand\bibfield[2]{#2}
\providecommand\bibinfo[2]{#2}
\providecommand\natexlab[1]{#1}
\providecommand\showeprint[2][]{arXiv:#2}

\bibitem[Adlakha et~al\mbox{.}(2023)]%
        {adlakha_evaluating_2023}
\bibfield{author}{\bibinfo{person}{Vaibhav Adlakha}, \bibinfo{person}{Parishad
  BehnamGhader}, \bibinfo{person}{Xing~Han Lu}, \bibinfo{person}{Nicholas
  Meade}, {and} \bibinfo{person}{Siva Reddy}.} \bibinfo{year}{2023}\natexlab{}.
\newblock \bibinfo{title}{Evaluating {Correctness} and {Faithfulness} of
  {Instruction}-{Following} {Models} for {Question} {Answering}}.
\newblock
\newblock
\urldef\tempurl%
\url{http://arxiv.org/abs/2307.16877}
\showURL{%
\tempurl}
\newblock
\shownote{arXiv:2307.16877 [cs]}.


\bibitem[Aleven et~al\mbox{.}(2023)]%
        {aleven_towards_2023}
\bibfield{author}{\bibinfo{person}{Vincent Aleven}, \bibinfo{person}{Richard
  Baraniuk}, \bibinfo{person}{Emma Brunskill}, \bibinfo{person}{Scott
  Crossley}, \bibinfo{person}{Dora Demszky}, \bibinfo{person}{Stephen
  Fancsali}, \bibinfo{person}{Shivang Gupta}, \bibinfo{person}{Kenneth
  Koedinger}, \bibinfo{person}{Chris Piech}, \bibinfo{person}{Steve Ritter},
  \bibinfo{person}{Danielle~R. Thomas}, \bibinfo{person}{Simon Woodhead}, {and}
  \bibinfo{person}{Wanli Xing}.} \bibinfo{year}{2023}\natexlab{}.
\newblock \showarticletitle{Towards the {Future} of {AI}-{Augmented} {Human}
  {Tutoring} in {Math} {Learning}}. In \bibinfo{booktitle}{\emph{Artificial
  {Intelligence} in {Education}. {Posters} and {Late} {Breaking} {Results},
  {Workshops} and {Tutorials}, {Industry} and {Innovation} {Tracks},
  {Practitioners}, {Doctoral} {Consortium} and {Blue} {Sky}}}
  \emph{(\bibinfo{series}{Communications in {Computer} and {Information}
  {Science}})}, \bibfield{editor}{\bibinfo{person}{Ning Wang},
  \bibinfo{person}{Genaro Rebolledo-Mendez}, \bibinfo{person}{Vania Dimitrova},
  \bibinfo{person}{Noboru Matsuda}, {and} \bibinfo{person}{Olga~C. Santos}}
  (Eds.). \bibinfo{publisher}{Springer Nature Switzerland},
  \bibinfo{address}{Cham}, \bibinfo{pages}{26--31}.
\newblock
\showISBNx{978-3-031-36336-8}
\urldef\tempurl%
\url{https://doi.org/10.1007/978-3-031-36336-8_3}
\showDOI{\tempurl}


\bibitem[Althammer et~al\mbox{.}(2022)]%
        {althammer_tripjudge_2022}
\bibfield{author}{\bibinfo{person}{Sophia Althammer},
  \bibinfo{person}{Sebastian Hofstätter}, \bibinfo{person}{Suzan Verberne},
  {and} \bibinfo{person}{Allan Hanbury}.} \bibinfo{year}{2022}\natexlab{}.
\newblock \showarticletitle{{TripJudge}: {A} {Relevance} {Judgement} {Test}
  {Collection} for {TripClick} {Health} {Retrieval}}. In
  \bibinfo{booktitle}{\emph{Proceedings of the 31st {ACM} {International}
  {Conference} on {Information} \& {Knowledge} {Management}}}.
  \bibinfo{publisher}{ACM}, \bibinfo{address}{Atlanta GA USA},
  \bibinfo{pages}{3801--3805}.
\newblock
\showISBNx{978-1-4503-9236-5}
\urldef\tempurl%
\url{https://doi.org/10.1145/3511808.3557714}
\showDOI{\tempurl}


\bibitem[Arroyo et~al\mbox{.}(2011)]%
        {arroyoUsingIntelligentTutor2011}
\bibfield{author}{\bibinfo{person}{Ivon Arroyo}, \bibinfo{person}{James~M
  Royer}, {and} \bibinfo{person}{Beverly~P Woolf}.}
  \bibinfo{year}{2011}\natexlab{}.
\newblock \showarticletitle{Using an Intelligent Tutor and Math Fluency
  Training to Improve Math Performance}.
\newblock \bibinfo{journal}{\emph{International Journal of Artificial
  Intelligence in Education}} \bibinfo{volume}{21}, \bibinfo{number}{1-2}
  (\bibinfo{year}{2011}), \bibinfo{pages}{135--152}.
\newblock
\showISSN{1560-4292}


\bibitem[Banawan et~al\mbox{.}(2022)]%
        {banawan_math_2022}
\bibfield{author}{\bibinfo{person}{Michelle Banawan}, \bibinfo{person}{Jinnie
  Shin}, \bibinfo{person}{Renu Balyan}, \bibinfo{person}{Walter~L. Leite},
  {and} \bibinfo{person}{Danielle~S. McNamara}.}
  \bibinfo{year}{2022}\natexlab{}.
\newblock \showarticletitle{Math {Discourse} {Linguistic} {Components}
  ({Cohesive} {Cues} within a {Math} {Discussion} {Board} {Discourse})}. In
  \bibinfo{booktitle}{\emph{Proceedings of the {Ninth} {ACM} {Conference} on
  {Learning} @ {Scale}}} \emph{(\bibinfo{series}{L@{S} '22})}.
  \bibinfo{publisher}{Association for Computing Machinery},
  \bibinfo{address}{New York, NY, USA}, \bibinfo{pages}{389--394}.
\newblock
\showISBNx{978-1-4503-9158-0}
\urldef\tempurl%
\url{https://doi.org/10.1145/3491140.3528320}
\showDOI{\tempurl}


\bibitem[Borgeaud et~al\mbox{.}(2022)]%
        {borgeaud_improving_2022}
\bibfield{author}{\bibinfo{person}{Sebastian Borgeaud}, \bibinfo{person}{Arthur
  Mensch}, \bibinfo{person}{Jordan Hoffmann}, \bibinfo{person}{Trevor Cai},
  \bibinfo{person}{Eliza Rutherford}, \bibinfo{person}{Katie Millican},
  \bibinfo{person}{George Bm Van~Den Driessche}, \bibinfo{person}{Jean-Baptiste
  Lespiau}, \bibinfo{person}{Bogdan Damoc}, \bibinfo{person}{Aidan Clark},
  \bibinfo{person}{Diego De~Las Casas}, \bibinfo{person}{Aurelia Guy},
  \bibinfo{person}{Jacob Menick}, \bibinfo{person}{Roman Ring},
  \bibinfo{person}{Tom Hennigan}, \bibinfo{person}{Saffron Huang},
  \bibinfo{person}{Loren Maggiore}, \bibinfo{person}{Chris Jones},
  \bibinfo{person}{Albin Cassirer}, \bibinfo{person}{Andy Brock},
  \bibinfo{person}{Michela Paganini}, \bibinfo{person}{Geoffrey Irving},
  \bibinfo{person}{Oriol Vinyals}, \bibinfo{person}{Simon Osindero},
  \bibinfo{person}{Karen Simonyan}, \bibinfo{person}{Jack Rae},
  \bibinfo{person}{Erich Elsen}, {and} \bibinfo{person}{Laurent Sifre}.}
  \bibinfo{year}{2022}\natexlab{}.
\newblock \showarticletitle{Improving {Language} {Models} by {Retrieving} from
  {Trillions} of {Tokens}}. In \bibinfo{booktitle}{\emph{Proceedings of the
  39th {International} {Conference} on {Machine} {Learning}}}.
  \bibinfo{publisher}{PMLR}, \bibinfo{pages}{2206--2240}.
\newblock
\urldef\tempurl%
\url{https://proceedings.mlr.press/v162/borgeaud22a.html}
\showURL{%
\tempurl}
\newblock
\shownote{ISSN: 2640-3498}.


\bibitem[Caines et~al\mbox{.}(2023)]%
        {caines_application_2023}
\bibfield{author}{\bibinfo{person}{Andrew Caines}, \bibinfo{person}{Luca
  Benedetto}, \bibinfo{person}{Shiva Taslimipoor}, \bibinfo{person}{Christopher
  Davis}, \bibinfo{person}{Yuan Gao}, \bibinfo{person}{Oeistein Andersen},
  \bibinfo{person}{Zheng Yuan}, \bibinfo{person}{Mark Elliott},
  \bibinfo{person}{Russell Moore}, \bibinfo{person}{Christopher Bryant},
  \bibinfo{person}{Marek Rei}, \bibinfo{person}{Helen Yannakoudakis},
  \bibinfo{person}{Andrew Mullooly}, \bibinfo{person}{Diane Nicholls}, {and}
  \bibinfo{person}{Paula Buttery}.} \bibinfo{year}{2023}\natexlab{}.
\newblock \bibinfo{title}{On the application of {Large} {Language} {Models} for
  language teaching and assessment technology}.
\newblock
\newblock
\urldef\tempurl%
\url{http://arxiv.org/abs/2307.08393}
\showURL{%
\tempurl}
\newblock
\shownote{arXiv:2307.08393 [cs]}.


\bibitem[Chase(2023)]%
        {chase_parent_2023}
\bibfield{author}{\bibinfo{person}{Harrison Chase}.}
  \bibinfo{year}{2023}\natexlab{}.
\newblock \bibinfo{title}{Parent {Document} {Retriever} - {LangChain}}.
\newblock
\newblock
\urldef\tempurl%
\url{https://python.langchain.com/docs/modules/data_connection/retrievers/parent_document_retriever}
\showURL{%
\tempurl}


\bibitem[Chen et~al\mbox{.}(2017)]%
        {chen_reading_2017}
\bibfield{author}{\bibinfo{person}{Danqi Chen}, \bibinfo{person}{Adam Fisch},
  \bibinfo{person}{Jason Weston}, {and} \bibinfo{person}{Antoine Bordes}.}
  \bibinfo{year}{2017}\natexlab{}.
\newblock \showarticletitle{Reading {Wikipedia} to {Answer} {Open}-{Domain}
  {Questions}}. In \bibinfo{booktitle}{\emph{Proceedings of the 55th {Annual}
  {Meeting} of the {Association} for {Computational} {Linguistics} ({Volume} 1:
  {Long} {Papers})}}, \bibfield{editor}{\bibinfo{person}{Regina Barzilay} {and}
  \bibinfo{person}{Min-Yen Kan}} (Eds.). \bibinfo{publisher}{Association for
  Computational Linguistics}, \bibinfo{address}{Vancouver, Canada},
  \bibinfo{pages}{1870--1879}.
\newblock
\urldef\tempurl%
\url{https://doi.org/10.18653/v1/P17-1171}
\showDOI{\tempurl}


\bibitem[Chiesurin et~al\mbox{.}(2023)]%
        {chiesurin_dangers_2023}
\bibfield{author}{\bibinfo{person}{Sabrina Chiesurin},
  \bibinfo{person}{Dimitris Dimakopoulos}, \bibinfo{person}{Marco~Antonio
  Sobrevilla~Cabezudo}, \bibinfo{person}{Arash Eshghi},
  \bibinfo{person}{Ioannis Papaioannou}, \bibinfo{person}{Verena Rieser}, {and}
  \bibinfo{person}{Ioannis Konstas}.} \bibinfo{year}{2023}\natexlab{}.
\newblock \showarticletitle{The {Dangers} of trusting {Stochastic} {Parrots}:
  {Faithfulness} and {Trust} in {Open}-domain {Conversational} {Question}
  {Answering}}. In \bibinfo{booktitle}{\emph{Findings of the {Association} for
  {Computational} {Linguistics}: {ACL} 2023}}. \bibinfo{publisher}{Association
  for Computational Linguistics}, \bibinfo{address}{Toronto, Canada},
  \bibinfo{pages}{947--959}.
\newblock
\urldef\tempurl%
\url{https://doi.org/10.18653/v1/2023.findings-acl.60}
\showDOI{\tempurl}


\bibitem[Cukurova et~al\mbox{.}(2022)]%
        {cukurovaLearningAnalyticsApproach2022}
\bibfield{author}{\bibinfo{person}{Mutlu Cukurova}, \bibinfo{person}{Madiha
  Khan-Galaria}, \bibinfo{person}{Eva Millán}, {and} \bibinfo{person}{Rose
  Luckin}.} \bibinfo{year}{2022}\natexlab{}.
\newblock \showarticletitle{A Learning Analytics Approach to Monitoring the
  Quality of Online One-to-One Tutoring}.
\newblock \bibinfo{journal}{\emph{Journal of Learning Analytics}}
  \bibinfo{volume}{9}, \bibinfo{number}{2} (\bibinfo{year}{2022}),
  \bibinfo{pages}{105--120}.
\newblock
\showISSN{1929-7750}


\bibitem[Dziri et~al\mbox{.}(2022)]%
        {dziri_faithdial_2022}
\bibfield{author}{\bibinfo{person}{Nouha Dziri}, \bibinfo{person}{Ehsan
  Kamalloo}, \bibinfo{person}{Sivan Milton}, \bibinfo{person}{Osmar Zaiane},
  \bibinfo{person}{Mo Yu}, \bibinfo{person}{Edoardo~M. Ponti}, {and}
  \bibinfo{person}{Siva Reddy}.} \bibinfo{year}{2022}\natexlab{}.
\newblock \bibinfo{title}{{FaithDial}: {A} {Faithful} {Benchmark} for
  {Information}-{Seeking} {Dialogue}}.
\newblock
\newblock
\urldef\tempurl%
\url{https://doi.org/10.48550/arXiv.2204.10757}
\showDOI{\tempurl}
\newblock
\shownote{arXiv:2204.10757 [cs]}.


\bibitem[Greene et~al\mbox{.}(2022)]%
        {greene_new_2022}
\bibfield{author}{\bibinfo{person}{Ryan Greene}, \bibinfo{person}{Ted Sanders},
  \bibinfo{person}{Lilian Weng}, {and} \bibinfo{person}{Arvind Neelakantan}.}
  \bibinfo{year}{2022}\natexlab{}.
\newblock \bibinfo{title}{New and improved embedding model}.
\newblock
\newblock
\urldef\tempurl%
\url{https://openai.com/blog/new-and-improved-embedding-model}
\showURL{%
\tempurl}


\bibitem[Guu et~al\mbox{.}(2020)]%
        {guu_realm_2020}
\bibfield{author}{\bibinfo{person}{Kelvin Guu}, \bibinfo{person}{Kenton Lee},
  \bibinfo{person}{Zora Tung}, \bibinfo{person}{Panupong Pasupat}, {and}
  \bibinfo{person}{Ming-Wei Chang}.} \bibinfo{year}{2020}\natexlab{}.
\newblock \showarticletitle{{REALM}: retrieval-augmented language model
  pre-training}. In \bibinfo{booktitle}{\emph{Proceedings of the 37th
  {International} {Conference} on {Machine} {Learning}}}
  \emph{(\bibinfo{series}{{ICML}'20}, Vol.~\bibinfo{volume}{119})}.
  \bibinfo{publisher}{JMLR.org}, \bibinfo{pages}{3929--3938}.
\newblock


\bibitem[Hofstätter et~al\mbox{.}(2020)]%
        {hofstatter_fine-grained_2020}
\bibfield{author}{\bibinfo{person}{Sebastian Hofstätter},
  \bibinfo{person}{Markus Zlabinger}, \bibinfo{person}{Mete Sertkan},
  \bibinfo{person}{Michael Schröder}, {and} \bibinfo{person}{Allan Hanbury}.}
  \bibinfo{year}{2020}\natexlab{}.
\newblock \showarticletitle{Fine-{Grained} {Relevance} {Annotations} for
  {Multi}-{Task} {Document} {Ranking} and {Question} {Answering}}. In
  \bibinfo{booktitle}{\emph{Proceedings of the 29th {ACM} {International}
  {Conference} on {Information} \& {Knowledge} {Management}}}
  \emph{(\bibinfo{series}{{CIKM} '20})}. \bibinfo{publisher}{Association for
  Computing Machinery}, \bibinfo{address}{New York, NY, USA},
  \bibinfo{pages}{3031--3038}.
\newblock
\showISBNx{978-1-4503-6859-9}
\urldef\tempurl%
\url{https://doi.org/10.1145/3340531.3412878}
\showDOI{\tempurl}


\bibitem[Holstein et~al\mbox{.}(2017)]%
        {holstein_intelligent_2017}
\bibfield{author}{\bibinfo{person}{Kenneth Holstein}, \bibinfo{person}{Bruce~M.
  McLaren}, {and} \bibinfo{person}{Vincent Aleven}.}
  \bibinfo{year}{2017}\natexlab{}.
\newblock \showarticletitle{Intelligent tutors as teachers' aides: exploring
  teacher needs for real-time analytics in blended classrooms}. In
  \bibinfo{booktitle}{\emph{Proceedings of the {Seventh} {International}
  {Learning} {Analytics} \& {Knowledge} {Conference}}}
  \emph{(\bibinfo{series}{{LAK} '17})}. \bibinfo{publisher}{Association for
  Computing Machinery}, \bibinfo{address}{New York, NY, USA},
  \bibinfo{pages}{257--266}.
\newblock
\showISBNx{978-1-4503-4870-6}
\urldef\tempurl%
\url{https://doi.org/10.1145/3027385.3027451}
\showDOI{\tempurl}


\bibitem[Hurrell(2021)]%
        {hurrellConceptualKnowledgeProcedural2021}
\bibfield{author}{\bibinfo{person}{Derek Hurrell}.}
  \bibinfo{year}{2021}\natexlab{}.
\newblock \showarticletitle{Conceptual Knowledge or Procedural Knowledge or
  Conceptual Knowledge and Procedural Knowledge: {{Why}} the Conjunction Is
  Important to Teachers}.
\newblock  \bibinfo{volume}{46}, \bibinfo{number}{2} (\bibinfo{year}{2021}),
  \bibinfo{pages}{57--71}.
\newblock
\showISSN{1835-517X}


\bibitem[Izacard et~al\mbox{.}(2022)]%
        {izacard_atlas_2022}
\bibfield{author}{\bibinfo{person}{Gautier Izacard}, \bibinfo{person}{Patrick
  Lewis}, \bibinfo{person}{Maria Lomeli}, \bibinfo{person}{Lucas Hosseini},
  \bibinfo{person}{Fabio Petroni}, \bibinfo{person}{Timo Schick},
  \bibinfo{person}{Jane Dwivedi-Yu}, \bibinfo{person}{Armand Joulin},
  \bibinfo{person}{Sebastian Riedel}, {and} \bibinfo{person}{Edouard Grave}.}
  \bibinfo{year}{2022}\natexlab{}.
\newblock \bibinfo{title}{Atlas: {Few}-shot {Learning} with {Retrieval}
  {Augmented} {Language} {Models}}.
\newblock
\newblock
\urldef\tempurl%
\url{https://doi.org/10.48550/arXiv.2208.03299}
\showDOI{\tempurl}
\newblock
\shownote{arXiv:2208.03299 [cs]}.


\bibitem[Ji et~al\mbox{.}(2023)]%
        {jiSurveyHallucinationNatural2023}
\bibfield{author}{\bibinfo{person}{Ziwei Ji}, \bibinfo{person}{Nayeon Lee},
  \bibinfo{person}{Rita Frieske}, \bibinfo{person}{Tiezheng Yu},
  \bibinfo{person}{Dan Su}, \bibinfo{person}{Yan Xu}, \bibinfo{person}{Etsuko
  Ishii}, \bibinfo{person}{Ye~Jin Bang}, \bibinfo{person}{Andrea Madotto},
  {and} \bibinfo{person}{Pascale Fung}.} \bibinfo{year}{2023}\natexlab{}.
\newblock \showarticletitle{Survey of Hallucination in Natural Language
  Generation}.
\newblock  \bibinfo{volume}{55}, \bibinfo{number}{12} (\bibinfo{year}{2023}),
  \bibinfo{pages}{1--38}.
\newblock
\showISSN{0360-0300}


\bibitem[Kasneci et~al\mbox{.}(2023)]%
        {kasneci_chatgpt_2023}
\bibfield{author}{\bibinfo{person}{Enkelejda Kasneci}, \bibinfo{person}{Kathrin
  Sessler}, \bibinfo{person}{Stefan Küchemann}, \bibinfo{person}{Maria
  Bannert}, \bibinfo{person}{Daryna Dementieva}, \bibinfo{person}{Frank
  Fischer}, \bibinfo{person}{Urs Gasser}, \bibinfo{person}{Georg Groh},
  \bibinfo{person}{Stephan Günnemann}, \bibinfo{person}{Eyke Hüllermeier},
  \bibinfo{person}{Stephan Krusche}, \bibinfo{person}{Gitta Kutyniok},
  \bibinfo{person}{Tilman Michaeli}, \bibinfo{person}{Claudia Nerdel},
  \bibinfo{person}{Jürgen Pfeffer}, \bibinfo{person}{Oleksandra Poquet},
  \bibinfo{person}{Michael Sailer}, \bibinfo{person}{Albrecht Schmidt},
  \bibinfo{person}{Tina Seidel}, \bibinfo{person}{Matthias Stadler},
  \bibinfo{person}{Jochen Weller}, \bibinfo{person}{Jochen Kuhn}, {and}
  \bibinfo{person}{Gjergji Kasneci}.} \bibinfo{year}{2023}\natexlab{}.
\newblock \showarticletitle{{ChatGPT} for good? {On} opportunities and
  challenges of large language models for education}.
\newblock \bibinfo{journal}{\emph{Learning and Individual Differences}}
  \bibinfo{volume}{103} (\bibinfo{date}{April} \bibinfo{year}{2023}),
  \bibinfo{pages}{102274}.
\newblock
\showISSN{1041-6080}
\urldef\tempurl%
\url{https://doi.org/10.1016/j.lindif.2023.102274}
\showDOI{\tempurl}


\bibitem[Kraft and Falken(2021)]%
        {kraftBlueprintScalingTutoring2021}
\bibfield{author}{\bibinfo{person}{Matthew~A Kraft} {and}
  \bibinfo{person}{Grace~T Falken}.} \bibinfo{year}{2021}\natexlab{}.
\newblock \showarticletitle{A Blueprint for Scaling Tutoring and Mentoring
  across Public Schools}.
\newblock   \bibinfo{volume}{7} (\bibinfo{year}{2021}),
  \bibinfo{pages}{23328584211042858}.
\newblock
\showISSN{2332-8584}


\bibitem[Lazaridou et~al\mbox{.}(2022)]%
        {lazaridou_internet-augmented_2022}
\bibfield{author}{\bibinfo{person}{Angeliki Lazaridou}, \bibinfo{person}{Elena
  Gribovskaya}, \bibinfo{person}{Wojciech~Jan Stokowiec}, {and}
  \bibinfo{person}{Nikolai Grigorev}.} \bibinfo{year}{2022}\natexlab{}.
\newblock \showarticletitle{Internet-augmented language models through few-shot
  prompting for open-domain question answering}.
\newblock  (\bibinfo{date}{Sept.} \bibinfo{year}{2022}).
\newblock
\urldef\tempurl%
\url{https://openreview.net/forum?id=hFCUPkSSRE}
\showURL{%
\tempurl}


\bibitem[Levonian et~al\mbox{.}(2023)]%
        {levonian_llm-math-education_2023}
\bibfield{author}{\bibinfo{person}{Zachary Levonian}, \bibinfo{person}{Owen
  Henkel}, {and} \bibinfo{person}{Bill Roberts}.}
  \bibinfo{year}{2023}\natexlab{}.
\newblock \bibinfo{title}{llm-math-education: {Retrieval} augmented generation
  for middle-school math question answering and hint generation}.
\newblock
\newblock
\urldef\tempurl%
\url{https://doi.org/10.5281/zenodo.8284412}
\showDOI{\tempurl}


\bibitem[Lewis et~al\mbox{.}(2020)]%
        {lewis_retrieval-augmented_2020}
\bibfield{author}{\bibinfo{person}{Patrick Lewis}, \bibinfo{person}{Ethan
  Perez}, \bibinfo{person}{Aleksandra Piktus}, \bibinfo{person}{Fabio Petroni},
  \bibinfo{person}{Vladimir Karpukhin}, \bibinfo{person}{Naman Goyal},
  \bibinfo{person}{Heinrich Küttler}, \bibinfo{person}{Mike Lewis},
  \bibinfo{person}{Wen-tau Yih}, \bibinfo{person}{Tim Rocktäschel},
  \bibinfo{person}{Sebastian Riedel}, {and} \bibinfo{person}{Douwe Kiela}.}
  \bibinfo{year}{2020}\natexlab{}.
\newblock \showarticletitle{Retrieval-augmented generation for
  knowledge-intensive {NLP} tasks}. In \bibinfo{booktitle}{\emph{Proceedings of
  the 34th {International} {Conference} on {Neural} {Information} {Processing}
  {Systems}}} \emph{(\bibinfo{series}{{NeurIPS}'20})}.
  \bibinfo{publisher}{Curran Associates Inc.}, \bibinfo{address}{Red Hook, NY,
  USA}, \bibinfo{pages}{9459--9474}.
\newblock
\showISBNx{978-1-71382-954-6}


\bibitem[Li and Xing(2021)]%
        {liNaturalLanguageGeneration2021}
\bibfield{author}{\bibinfo{person}{Chenglu Li} {and} \bibinfo{person}{Wanli
  Xing}.} \bibinfo{year}{2021}\natexlab{}.
\newblock \showarticletitle{Natural Language Generation Using Deep Learning to
  Support {{MOOC}} Learners}.
\newblock   \bibinfo{volume}{31} (\bibinfo{year}{2021}),
  \bibinfo{pages}{186--214}.
\newblock
\showISSN{1560-4292}


\bibitem[Lin et~al\mbox{.}(2021)]%
        {lin_pyserini_2021}
\bibfield{author}{\bibinfo{person}{Jimmy Lin}, \bibinfo{person}{Xueguang Ma},
  \bibinfo{person}{Sheng-Chieh Lin}, \bibinfo{person}{Jheng-Hong Yang},
  \bibinfo{person}{Ronak Pradeep}, {and} \bibinfo{person}{Rodrigo Nogueira}.}
  \bibinfo{year}{2021}\natexlab{}.
\newblock \showarticletitle{Pyserini: {A} {Python} {Toolkit} for {Reproducible}
  {Information} {Retrieval} {Research} with {Sparse} and {Dense}
  {Representations}}. In \bibinfo{booktitle}{\emph{Proceedings of the 44th
  {International} {ACM} {SIGIR} {Conference} on {Research} and {Development} in
  {Information} {Retrieval}}} \emph{(\bibinfo{series}{{SIGIR} '21})}.
  \bibinfo{publisher}{Association for Computing Machinery},
  \bibinfo{address}{New York, NY, USA}, \bibinfo{pages}{2356--2362}.
\newblock
\showISBNx{978-1-4503-8037-9}
\urldef\tempurl%
\url{https://doi.org/10.1145/3404835.3463238}
\showDOI{\tempurl}


\bibitem[Lu et~al\mbox{.}(2023)]%
        {lu_are_2023}
\bibfield{author}{\bibinfo{person}{Sheng Lu}, \bibinfo{person}{Irina
  Bigoulaeva}, \bibinfo{person}{Rachneet Sachdeva},
  \bibinfo{person}{Harish~Tayyar Madabushi}, {and} \bibinfo{person}{Iryna
  Gurevych}.} \bibinfo{year}{2023}\natexlab{}.
\newblock \bibinfo{title}{Are {Emergent} {Abilities} in {Large} {Language}
  {Models} just {In}-{Context} {Learning}?}
\newblock
\newblock
\urldef\tempurl%
\url{https://doi.org/10.48550/arXiv.2309.01809}
\showDOI{\tempurl}
\newblock
\shownote{arXiv:2309.01809 [cs]}.


\bibitem[Marecek et~al\mbox{.}(2020)]%
        {marecek_prealgebra_2020}
\bibfield{author}{\bibinfo{person}{Lynn Marecek}, \bibinfo{person}{MaryAnne
  Anthony-Smith}, {and} \bibinfo{person}{Andrea Honeycutt~Mathis}.}
  \bibinfo{year}{2020}\natexlab{}.
\newblock \bibinfo{booktitle}{\emph{Prealgebra} (\bibinfo{edition}{2} ed.)}.
\newblock
\showISBNx{978-1-951693-19-0}
\urldef\tempurl%
\url{https://openstax.org/details/books/prealgebra-2e}
\showURL{%
\tempurl}


\bibitem[Mialon et~al\mbox{.}(2023)]%
        {mialon_augmented_2023}
\bibfield{author}{\bibinfo{person}{Grégoire Mialon}, \bibinfo{person}{Roberto
  Dessì}, \bibinfo{person}{Maria Lomeli}, \bibinfo{person}{Christoforos
  Nalmpantis}, \bibinfo{person}{Ram Pasunuru}, \bibinfo{person}{Roberta
  Raileanu}, \bibinfo{person}{Baptiste Rozière}, \bibinfo{person}{Timo
  Schick}, \bibinfo{person}{Jane Dwivedi-Yu}, \bibinfo{person}{Asli
  Celikyilmaz}, \bibinfo{person}{Edouard Grave}, \bibinfo{person}{Yann LeCun},
  {and} \bibinfo{person}{Thomas Scialom}.} \bibinfo{year}{2023}\natexlab{}.
\newblock \bibinfo{title}{Augmented {Language} {Models}: a {Survey}}.
\newblock
\newblock
\urldef\tempurl%
\url{http://arxiv.org/abs/2302.07842}
\showURL{%
\tempurl}
\newblock
\shownote{arXiv:2302.07842 [cs]}.


\bibitem[Mishra et~al\mbox{.}(2022)]%
        {mishra_reframing_2022}
\bibfield{author}{\bibinfo{person}{Swaroop Mishra}, \bibinfo{person}{Daniel
  Khashabi}, \bibinfo{person}{Chitta Baral}, \bibinfo{person}{Yejin Choi},
  {and} \bibinfo{person}{Hannaneh Hajishirzi}.}
  \bibinfo{year}{2022}\natexlab{}.
\newblock \showarticletitle{Reframing {Instructional} {Prompts} to {GPTk}'s
  {Language}}. In \bibinfo{booktitle}{\emph{Findings of the {Association} for
  {Computational} {Linguistics}: {ACL} 2022}}. \bibinfo{publisher}{Association
  for Computational Linguistics}, \bibinfo{address}{Dublin, Ireland},
  \bibinfo{pages}{589--612}.
\newblock
\urldef\tempurl%
\url{https://doi.org/10.18653/v1/2022.findings-acl.50}
\showDOI{\tempurl}


\bibitem[Moschkovich(2015)]%
        {moschkovichScaffoldingStudentParticipation2015}
\bibfield{author}{\bibinfo{person}{Judit~N Moschkovich}.}
  \bibinfo{year}{2015}\natexlab{}.
\newblock \showarticletitle{Scaffolding Student Participation in Mathematical
  Practices}.
\newblock   \bibinfo{volume}{47} (\bibinfo{year}{2015}),
  \bibinfo{pages}{1067--1078}.
\newblock
\showISSN{1863-9690}


\bibitem[Muennighoff et~al\mbox{.}(2023)]%
        {muennighoff_mteb_2023}
\bibfield{author}{\bibinfo{person}{Niklas Muennighoff},
  \bibinfo{person}{Nouamane Tazi}, \bibinfo{person}{Loïc Magne}, {and}
  \bibinfo{person}{Nils Reimers}.} \bibinfo{year}{2023}\natexlab{}.
\newblock \bibinfo{title}{{MTEB}: {Massive} {Text} {Embedding} {Benchmark}}.
\newblock
\newblock
\urldef\tempurl%
\url{https://doi.org/10.48550/arXiv.2210.07316}
\showDOI{\tempurl}
\newblock
\shownote{arXiv:2210.07316 [cs]}.


\bibitem[NAEP(2022)]%
        {naepNAEPMathematicsNational}
\bibfield{author}{\bibinfo{person}{NAEP}.} \bibinfo{year}{2022}\natexlab{}.
\newblock \bibinfo{booktitle}{\emph{{{NAEP Mathematics}}: {{National Average
  Scores}}}}.
\newblock
\urldef\tempurl%
\url{https://www.nationsreportcard.gov/mathematics/nation/scores/?grade=8}
\showURL{%
\tempurl}


\bibitem[Nickow et~al\mbox{.}(2020)]%
        {nickow_impressive_2020}
\bibfield{author}{\bibinfo{person}{Andre Nickow}, \bibinfo{person}{Philip
  Oreopoulos}, {and} \bibinfo{person}{Vincent Quan}.}
  \bibinfo{year}{2020}\natexlab{}.
\newblock \showarticletitle{The impressive effects of tutoring on prek-12
  learning: {A} systematic review and meta-analysis of the experimental
  evidence}.
\newblock  (\bibinfo{year}{2020}).
\newblock
\newblock
\shownote{Publisher: National Bureau of Economic Research}.


\bibitem[Nye et~al\mbox{.}(2023)]%
        {nye_generative_2023}
\bibfield{author}{\bibinfo{person}{Benjamin~D Nye}, \bibinfo{person}{Dillon
  Mee}, {and} \bibinfo{person}{Mark~G Core}.} \bibinfo{year}{2023}\natexlab{}.
\newblock \showarticletitle{Generative {Large} {Language} {Models} for
  {Dialog}-{Based} {Tutoring}: {An} {Early} {Consideration} of {Opportunities}
  and {Concerns}}. \bibinfo{address}{Tokyo, Japan}.
\newblock
\urldef\tempurl%
\url{https://ceur-ws.org/Vol-3487/paper4.pdf}
\showURL{%
\tempurl}


\bibitem[Peng et~al\mbox{.}(2023)]%
        {peng_check_2023}
\bibfield{author}{\bibinfo{person}{Baolin Peng}, \bibinfo{person}{Michel
  Galley}, \bibinfo{person}{Pengcheng He}, \bibinfo{person}{Hao Cheng},
  \bibinfo{person}{Yujia Xie}, \bibinfo{person}{Yu Hu},
  \bibinfo{person}{Qiuyuan Huang}, \bibinfo{person}{Lars Liden},
  \bibinfo{person}{Zhou Yu}, \bibinfo{person}{Weizhu Chen}, {and}
  \bibinfo{person}{Jianfeng Gao}.} \bibinfo{year}{2023}\natexlab{}.
\newblock \bibinfo{title}{Check {Your} {Facts} and {Try} {Again}: {Improving}
  {Large} {Language} {Models} with {External} {Knowledge} and {Automated}
  {Feedback}}.
\newblock
\newblock
\urldef\tempurl%
\url{http://arxiv.org/abs/2302.12813}
\showURL{%
\tempurl}
\newblock
\shownote{arXiv:2302.12813 [cs]}.


\bibitem[Psotka et~al\mbox{.}(1988)]%
        {psotka_intelligent_1988}
\bibfield{editor}{\bibinfo{person}{Joseph Psotka}, \bibinfo{person}{L.~Dan
  Massey}, {and} \bibinfo{person}{Sharon~A. Mutter}} (Eds.).
  \bibinfo{year}{1988}\natexlab{}.
\newblock \bibinfo{booktitle}{\emph{Intelligent tutoring systems: {Lessons}
  learned}}.
\newblock \bibinfo{publisher}{Lawrence Erlbaum Associates, Inc},
  \bibinfo{address}{Hillsdale, NJ, US}.
\newblock
\showISBNx{978-0-8058-0023-4 978-0-8058-0192-7}
\newblock
\shownote{Pages: xxii, 552}.


\bibitem[Rajpurkar et~al\mbox{.}(2016)]%
        {rajpurkar_squad_2016}
\bibfield{author}{\bibinfo{person}{Pranav Rajpurkar}, \bibinfo{person}{Jian
  Zhang}, \bibinfo{person}{Konstantin Lopyrev}, {and} \bibinfo{person}{Percy
  Liang}.} \bibinfo{year}{2016}\natexlab{}.
\newblock \bibinfo{title}{{SQuAD}: 100,000+ {Questions} for {Machine}
  {Comprehension} of {Text}}.
\newblock
\newblock
\urldef\tempurl%
\url{http://arxiv.org/abs/1606.05250}
\showURL{%
\tempurl}
\newblock
\shownote{arXiv:1606.05250 [cs]}.


\bibitem[Ritter et~al\mbox{.}(2007)]%
        {ritterCognitiveTutorApplied2007}
\bibfield{author}{\bibinfo{person}{Steven Ritter}, \bibinfo{person}{John~R
  Anderson}, \bibinfo{person}{Kenneth~R Koedinger}, {and}
  \bibinfo{person}{Albert Corbett}.} \bibinfo{year}{2007}\natexlab{}.
\newblock \showarticletitle{Cognitive {{Tutor}}: {{Applied}} Research in
  Mathematics Education}.
\newblock   \bibinfo{volume}{14} (\bibinfo{year}{2007}),
  \bibinfo{pages}{249--255}.
\newblock
\showISSN{1069-9384}


\bibitem[Rittle-Johnson et~al\mbox{.}(2015)]%
        {rittle-johnsonNotOnewayStreet2015}
\bibfield{author}{\bibinfo{person}{Bethany Rittle-Johnson},
  \bibinfo{person}{Michael Schneider}, {and} \bibinfo{person}{Jon~R Star}.}
  \bibinfo{year}{2015}\natexlab{}.
\newblock \showarticletitle{Not a One-Way Street: {{Bidirectional}} Relations
  between Procedural and Conceptual Knowledge of Mathematics}.
\newblock   \bibinfo{volume}{27} (\bibinfo{year}{2015}),
  \bibinfo{pages}{587--597}.
\newblock
\showISSN{1040-726X}


\bibitem[Sedlmeier(2001)]%
        {sedlmeier_intelligent_2001}
\bibfield{author}{\bibinfo{person}{P. Sedlmeier}.}
  \bibinfo{year}{2001}\natexlab{}.
\newblock \showarticletitle{Intelligent {Tutoring} {Systems}}.
\newblock In \bibinfo{booktitle}{\emph{International {Encyclopedia} of the
  {Social} \& {Behavioral} {Sciences}}},
  \bibfield{editor}{\bibinfo{person}{Neil~J. Smelser} {and}
  \bibinfo{person}{Paul~B. Baltes}} (Eds.). \bibinfo{publisher}{Pergamon},
  \bibinfo{address}{Oxford}, \bibinfo{pages}{7674--7678}.
\newblock
\showISBNx{978-0-08-043076-8}
\urldef\tempurl%
\url{https://doi.org/10.1016/B0-08-043076-7/01618-1}
\showDOI{\tempurl}


\bibitem[Sellam et~al\mbox{.}(2020)]%
        {sellam_bleurt_2020}
\bibfield{author}{\bibinfo{person}{Thibault Sellam}, \bibinfo{person}{Dipanjan
  Das}, {and} \bibinfo{person}{Ankur Parikh}.} \bibinfo{year}{2020}\natexlab{}.
\newblock \showarticletitle{{BLEURT}: {Learning} {Robust} {Metrics} for {Text}
  {Generation}}. In \bibinfo{booktitle}{\emph{Proceedings of the 58th {Annual}
  {Meeting} of the {Association} for {Computational} {Linguistics}}}.
  \bibinfo{publisher}{Association for Computational Linguistics},
  \bibinfo{address}{Online}, \bibinfo{pages}{7881--7892}.
\newblock
\urldef\tempurl%
\url{https://doi.org/10.18653/v1/2020.acl-main.704}
\showDOI{\tempurl}


\bibitem[Shen et~al\mbox{.}(2021)]%
        {shenMathBERTPretrainedLanguage2021}
\bibfield{author}{\bibinfo{person}{Jia~Tracy Shen}, \bibinfo{person}{Michiharu
  Yamashita}, \bibinfo{person}{Ethan Prihar}, \bibinfo{person}{Neil Heffernan},
  \bibinfo{person}{Xintao Wu}, \bibinfo{person}{Ben Graff}, {and}
  \bibinfo{person}{Dongwon Lee}.} \bibinfo{year}{2021}\natexlab{}.
\newblock \showarticletitle{{{MathBERT}}: {{A Pre-trained Language Model}} for
  {{General NLP Tasks}} in {{Mathematics Education}}}.
\newblock


\bibitem[Sonkar et~al\mbox{.}(2023)]%
        {sonkar_class_2023}
\bibfield{author}{\bibinfo{person}{Shashank Sonkar}, \bibinfo{person}{Lucy
  Liu}, \bibinfo{person}{Debshila~Basu Mallick}, {and}
  \bibinfo{person}{Richard~G. Baraniuk}.} \bibinfo{year}{2023}\natexlab{}.
\newblock \bibinfo{title}{{CLASS} {Meet} {SPOCK}: {An} {Education} {Tutoring}
  {Chatbot} based on {Learning} {Science} {Principles}}.
\newblock
\newblock
\urldef\tempurl%
\url{http://arxiv.org/abs/2305.13272}
\showURL{%
\tempurl}
\newblock
\shownote{arXiv:2305.13272 [cs]}.


\bibitem[Sottilare et~al\mbox{.}(2014)]%
        {sottilare_design_2014}
\bibfield{author}{\bibinfo{person}{Robert~A Sottilare}, \bibinfo{person}{Arthur
  Graesser}, \bibinfo{person}{Xiangen Hu}, {and} \bibinfo{person}{Benjamin~S
  Goldberg}.} \bibinfo{year}{2014}\natexlab{}.
\newblock \bibinfo{booktitle}{\emph{Design {Recommendations} for {Intelligent}
  {Tutoring} {Systems}. {Volume} 2: {Instructional} {Management}}}.
\newblock \bibinfo{type}{{T}echnical {R}eport}.
  \bibinfo{institution}{UNIVERSITY OF SOUTHERN CALIFORNIA LOS ANGELES}.
  \bibinfo{pages}{427} pages.
\newblock
\urldef\tempurl%
\url{https://apps.dtic.mil/sti/citations/AD1158927}
\showURL{%
\tempurl}
\newblock
\shownote{Section: Technical Reports}.


\bibitem[Sung et~al\mbox{.}(2021)]%
        {sungHowDoesAugmented2021}
\bibfield{author}{\bibinfo{person}{Shannon~H Sung}, \bibinfo{person}{Chenglu
  Li}, \bibinfo{person}{Guanhua Chen}, \bibinfo{person}{Xudong Huang},
  \bibinfo{person}{Charles Xie}, \bibinfo{person}{Joyce Massicotte}, {and}
  \bibinfo{person}{Ji Shen}.} \bibinfo{year}{2021}\natexlab{}.
\newblock \showarticletitle{How Does Augmented Observation Facilitate
  Multimodal Representational Thinking? {{Applying}} Deep Learning to Decode
  Complex Student Construct}.
\newblock   \bibinfo{volume}{30} (\bibinfo{year}{2021}),
  \bibinfo{pages}{210--226}.
\newblock
\showISSN{1059-0145}


\bibitem[Upadhyay et~al\mbox{.}(2023)]%
        {upadhyay_improving_2023}
\bibfield{author}{\bibinfo{person}{Shriyash Upadhyay}, \bibinfo{person}{Etan
  Ginsberg}, {and} \bibinfo{person}{Chris Callison-Burch}.}
  \bibinfo{year}{2023}\natexlab{}.
\newblock \showarticletitle{Improving {Mathematics} {Tutoring} {With} {A}
  {Code} {Scratchpad}}. In \bibinfo{booktitle}{\emph{Proceedings of the 18th
  {Workshop} on {Innovative} {Use} of {NLP} for {Building} {Educational}
  {Applications} ({BEA} 2023)}}. \bibinfo{publisher}{Association for
  Computational Linguistics}, \bibinfo{address}{Toronto, Canada},
  \bibinfo{pages}{20--28}.
\newblock
\urldef\tempurl%
\url{https://doi.org/10.18653/v1/2023.bea-1.2}
\showDOI{\tempurl}


\bibitem[VanLehn(2006)]%
        {vanlehn_behavior_2006}
\bibfield{author}{\bibinfo{person}{Kurt VanLehn}.}
  \bibinfo{year}{2006}\natexlab{}.
\newblock \showarticletitle{The {Behavior} of {Tutoring} {Systems}}.
\newblock \bibinfo{journal}{\emph{International Journal of Artificial
  Intelligence in Education}} \bibinfo{volume}{16}, \bibinfo{number}{3}
  (\bibinfo{date}{Aug.} \bibinfo{year}{2006}), \bibinfo{pages}{227--265}.
\newblock
\showISSN{1560-4292}


\bibitem[VanLehn(2011)]%
        {vanlehn_relative_2011}
\bibfield{author}{\bibinfo{person}{Kurt VanLehn}.}
  \bibinfo{year}{2011}\natexlab{}.
\newblock \showarticletitle{The relative effectiveness of human tutoring,
  intelligent tutoring systems, and other tutoring systems}.
\newblock \bibinfo{journal}{\emph{Educational Psychologist}}
  \bibinfo{volume}{46}, \bibinfo{number}{4} (\bibinfo{year}{2011}),
  \bibinfo{pages}{197--221}.
\newblock
\showISSN{1532-6985}
\urldef\tempurl%
\url{https://doi.org/10.1080/00461520.2011.611369}
\showDOI{\tempurl}
\newblock
\shownote{Place: United Kingdom Publisher: Taylor \& Francis}.


\bibitem[Wang et~al\mbox{.}(2023)]%
        {wang_mint_2023}
\bibfield{author}{\bibinfo{person}{Xingyao Wang}, \bibinfo{person}{Zihan Wang},
  \bibinfo{person}{Jiateng Liu}, \bibinfo{person}{Yangyi Chen},
  \bibinfo{person}{Lifan Yuan}, \bibinfo{person}{Hao Peng}, {and}
  \bibinfo{person}{Heng Ji}.} \bibinfo{year}{2023}\natexlab{}.
\newblock \bibinfo{title}{{MINT}: {Evaluating} {LLMs} in {Multi}-turn
  {Interaction} with {Tools} and {Language} {Feedback}}.
\newblock
\newblock
\urldef\tempurl%
\url{https://doi.org/10.48550/arXiv.2309.10691}
\showDOI{\tempurl}
\newblock
\shownote{arXiv:2309.10691 [cs]}.


\bibitem[Yang et~al\mbox{.}(2023b)]%
        {yang_large_2023}
\bibfield{author}{\bibinfo{person}{Chengrun Yang}, \bibinfo{person}{Xuezhi
  Wang}, \bibinfo{person}{Yifeng Lu}, \bibinfo{person}{Hanxiao Liu},
  \bibinfo{person}{Quoc~V. Le}, \bibinfo{person}{Denny Zhou}, {and}
  \bibinfo{person}{Xinyun Chen}.} \bibinfo{year}{2023}\natexlab{b}.
\newblock \bibinfo{title}{Large {Language} {Models} as {Optimizers}}.
\newblock
\newblock
\urldef\tempurl%
\url{http://arxiv.org/abs/2309.03409}
\showURL{%
\tempurl}
\newblock
\shownote{arXiv:2309.03409 [cs]}.


\bibitem[Yang et~al\mbox{.}(2023a)]%
        {yang_leandojo_2023}
\bibfield{author}{\bibinfo{person}{Kaiyu Yang}, \bibinfo{person}{Aidan~M.
  Swope}, \bibinfo{person}{Alex Gu}, \bibinfo{person}{Rahul Chalamala},
  \bibinfo{person}{Peiyang Song}, \bibinfo{person}{Shixing Yu},
  \bibinfo{person}{Saad Godil}, \bibinfo{person}{Ryan Prenger}, {and}
  \bibinfo{person}{Anima Anandkumar}.} \bibinfo{year}{2023}\natexlab{a}.
\newblock \bibinfo{title}{{LeanDojo}: {Theorem} {Proving} with
  {Retrieval}-{Augmented} {Language} {Models}}.
\newblock
\newblock
\urldef\tempurl%
\url{http://arxiv.org/abs/2306.15626}
\showURL{%
\tempurl}
\newblock
\shownote{arXiv:2306.15626 [cs, stat]}.


\bibitem[Yang et~al\mbox{.}(2021)]%
        {yang_can_2021}
\bibfield{author}{\bibinfo{person}{Kexin~Bella Yang}, \bibinfo{person}{Tomohiro
  Nagashima}, \bibinfo{person}{Junhui Yao}, \bibinfo{person}{Joseph~Jay
  Williams}, \bibinfo{person}{Kenneth Holstein}, {and} \bibinfo{person}{Vincent
  Aleven}.} \bibinfo{year}{2021}\natexlab{}.
\newblock \showarticletitle{Can {Crowds} {Customize} {Instructional}
  {Materials} with {Minimal} {Expert} {Guidance}? {Exploring} {Teacher}-guided
  {Crowdsourcing} for {Improving} {Hints} in an {AI}-based {Tutor}}.
\newblock \bibinfo{journal}{\emph{Proceedings of the ACM on Human-Computer
  Interaction}} \bibinfo{volume}{5}, \bibinfo{number}{CSCW1}
  (\bibinfo{date}{April} \bibinfo{year}{2021}), \bibinfo{pages}{119:1--119:24}.
\newblock
\urldef\tempurl%
\url{https://doi.org/10.1145/3449193}
\showDOI{\tempurl}


\bibitem[Yang et~al\mbox{.}(2019)]%
        {yang_critically_2019}
\bibfield{author}{\bibinfo{person}{Wei Yang}, \bibinfo{person}{Kuang Lu},
  \bibinfo{person}{Peilin Yang}, {and} \bibinfo{person}{Jimmy Lin}.}
  \bibinfo{year}{2019}\natexlab{}.
\newblock \showarticletitle{Critically {Examining} the "{Neural} {Hype}":
  {Weak} {Baselines} and the {Additivity} of {Effectiveness} {Gains} from
  {Neural} {Ranking} {Models}}. In \bibinfo{booktitle}{\emph{Proceedings of the
  42nd {International} {ACM} {SIGIR} {Conference} on {Research} and
  {Development} in {Information} {Retrieval}}}
  \emph{(\bibinfo{series}{{SIGIR}'19})}. \bibinfo{publisher}{Association for
  Computing Machinery}, \bibinfo{address}{New York, NY, USA},
  \bibinfo{pages}{1129--1132}.
\newblock
\showISBNx{978-1-4503-6172-9}
\urldef\tempurl%
\url{https://doi.org/10.1145/3331184.3331340}
\showDOI{\tempurl}


\bibitem[Zamani et~al\mbox{.}(2022)]%
        {zamani_retrieval-enhanced_2022}
\bibfield{author}{\bibinfo{person}{Hamed Zamani}, \bibinfo{person}{Fernando
  Diaz}, \bibinfo{person}{Mostafa Dehghani}, \bibinfo{person}{Donald Metzler},
  {and} \bibinfo{person}{Michael Bendersky}.} \bibinfo{year}{2022}\natexlab{}.
\newblock \showarticletitle{Retrieval-{Enhanced} {Machine} {Learning}}. In
  \bibinfo{booktitle}{\emph{Proceedings of the 45th {International} {ACM}
  {SIGIR} {Conference} on {Research} and {Development} in {Information}
  {Retrieval}}} \emph{(\bibinfo{series}{{SIGIR} '22})}.
  \bibinfo{publisher}{Association for Computing Machinery},
  \bibinfo{address}{New York, NY, USA}, \bibinfo{pages}{2875--2886}.
\newblock
\showISBNx{978-1-4503-8732-3}
\urldef\tempurl%
\url{https://doi.org/10.1145/3477495.3531722}
\showDOI{\tempurl}


\bibitem[Zhang et~al\mbox{.}(2020)]%
        {zhang_bertscore_2020}
\bibfield{author}{\bibinfo{person}{Tianyi Zhang}, \bibinfo{person}{Varsha
  Kishore}, \bibinfo{person}{Felix Wu}, \bibinfo{person}{Kilian~Q. Weinberger},
  {and} \bibinfo{person}{Yoav Artzi}.} \bibinfo{year}{2020}\natexlab{}.
\newblock \showarticletitle{{BERTScore}: {Evaluating} {Text} {Generation} with
  {BERT}}.
\newblock
\urldef\tempurl%
\url{https://openreview.net/forum?id=SkeHuCVFDr}
\showURL{%
\tempurl}


\end{thebibliography}

\appendix
\section{Implementation details}
\label{app:sec:implementation}

We opted to use GPT-3.5 rather than GPT-4 because it reflects a more realistic cost trade-off for the Rori ITS system we are researching.
At the time of the study, GPT-3.5 had a context window of 4K tokens; we used up to 3K tokens for document retrieval. The median chapter and sub-section has 5,050 and 185 tokens respectively.
We chose dense retrieval both for its popularity in RAG implementations and its dominance on a related retrieval task (not reported here) compared to a strong sparse-retrieval baseline: Pyserini's BM25 implementation~\cite{lin_pyserini_2021,yang_critically_2019}).

\section{Prompts}
\label{app:sec:prompts}

Prompts used in the various guidance conditions.
``\{openstax\_text\}'' is replaced with the retrieved text.
The None, Low, and High guidance prompts are provided as system prompts, with the student question provided in a separate user prompt. The IR prompt is provided as a user prompt with ``\{query\}'' replaced by the student question.

\subsection{No guidance (None) prompt}

You are going to act as a mathematics tutor for a 13 year old student who is in grade 8 or 9 and lives in Ghana.

You will be encouraging and factual.

Prefer simple, short responses.

If the student says something inappropriate or off topic you will say you can only focus on mathematics and ask them if they have any math-related follow-up questions.

\subsection{Low guidance (Low) prompt}

You are going to act as a mathematics tutor for a 13 year old student who is in grade 8 or 9 and lives in Ghana.

You will be encouraging and factual.

Only if it is relevant, examples and language from the section below may be helpful to format your response:

===

\{openstax\_text\}

===

Prefer simple, short responses.

If the student says something inappropriate or off topic you will say you can only focus on mathematics and ask them if they have any math-related follow-up questions.

\subsection{High guidance (High) prompt}

You are going to act as a mathematics tutor for a 13 year old student who is in grade 8 or 9 and lives in Ghana.

You will be encouraging and factual.

Use examples and language from the section below to format your response:

===

\{openstax\_text\}

===

Prefer simple, short responses.

If the student says something inappropriate or off topic you will say you can only focus on mathematics and ask them if they have any math-related follow-up questions.

\subsection{Information Retrieval (IR) prompt}

Given a middle-school math student's question, you will identify the most relevant section from a textbook.

Student question: \{query\}

Repeat the student's question and then repeat in full the most relevant paragraph from my math textbook. If none of them seem relevant, take a deep breath and output the most relevant. Don't say anything else.

Textbook paragraphs:

\{openstax\_text\}

\section{Ranking \& Groundedness Survey}
\label{app:sec:ranking_survey}

Queries were split into four Qualtrics surveys; three surveys had 15 questions while the fourth had 6 questions. This section gives the exact survey text presented to respondents.
30 queries were annotated three times and the remaining 41 were annotated four times.
Table~\ref{app:tab:annotator_counts} shows per-annotator counts.

\begin{table}
\caption{Number of unique queries annotated by each survey respondent. }
\centering
\begin{tabular}{lr}
\toprule
Annotator & Query Count \\
\midrule
A1 & 30 \\
A2 & 30 \\
A3 & 21 \\
A4 & 21 \\
A5 & 21 \\
A6 & 15 \\
A7 & 15 \\
A8 & 15 \\
A9 & 6 \\
\bottomrule
\end{tabular}
\label{app:tab:annotator_counts}
\end{table}

\subsection{Intro page}

This survey will consist of 15 questions. Your progress will save after each question.

Who are you? (Annotator name) \underline{\hspace{3cm}}

\subsection{Query page}

(Survey format note: this page is repeated once for each query in the survey.)

\subsubsection{Ranking question}
Rank these three responses from best to worst response. Consider if the response answers the question and is factually correct.

Student's question:

\{query\}

\begin{tabular}{lccc}
 & 1 & 2 & 3 \\
\{response1\} & ◯ & ◯ & ◯ \\
\{response2\} & ◯ & ◯ & ◯ \\
\{response3\} & ◯ & ◯ & ◯ \\
\end{tabular}

\subsubsection{Groundedness question}

For each response, does the response or a paraphrase of the response appear anywhere in the following document?

Note: "First response" refers to the first response in the order they appear above, NOT the document you ranked as "1".

The document:

\{openstax\_text\}

None: The response, even paraphrased, does not appear anywhere in the document.

Partial: Part of the response (or a paraphrase of the response) appears in the document.

Perfect: The response (or a paraphrase of the response) appears in the document.

\begin{tabular}{lccc}
 & None & Partial & Perfect \\
First response & ◯ & ◯ & ◯ \\
Second response & ◯ & ◯ & ◯ \\
Third response & ◯ & ◯ & ◯ \\
\end{tabular}

\subsubsection{Qualitative observation question}

Notes/observations, if you want to flag something for later discussion with other annotators or if you spot a survey problem: \underline{\hspace{3cm}}


\section{Relevance Survey}
\label{app:sec:relevance_survey}

Three respondents (A1, A6, and A10) each independently annotated the 51 queries for relevance in separate tabs of a Google Sheet.

\subsection{Annotator instructions}

Each row contains a middle-school student's question (called the \textbf{query}) and an excerpt from a math textbook (called the \textbf{document}).
Your task is to decide if the document is relevant to the query.

Your options are:

\textbf{Wrong:} The document has nothing to do with the query, and does not help in any way to answer it.

\textbf{Topic:} The document talks about the general area or topic of a query, might provide some background info, but ultimately does not answer it.

\textbf{Partial:} The document contains a partial answer, but you think there should be more to it.

\textbf{Perfect:} The document contains a full answer: easy to understand and it directly answers the question in full.

For readability, I bullet-pointed the paragraphs within each document. It's okay if only one paragraph within the document is relevant: if any paragraph within the document contains a full (or partial) answer, that is sufficient.

Each annotator has their own sheet within this workbook; annotate only within your own sheet, and don't look at others annotations.

\subsection{Spreadsheet tab}

The annotation sheet had the following columns: query, document, relevance

\end{document}